# Antibody Foundational Model : Ab-RoBERTa

Eunna Huh[1]     Hyeonsu Lee[1]     Hyunjin Shin[1*]

[1] Mogam institute for biomedical research, South Korea
[*] Corresponding author: hyunjin.shin@mogam.re.kr

## Abstract

With the growing prominence of antibody-based therapeutics, antibody engineering has gained increasing attention as a critical area of research and development. Recent progress in transformer-based protein large language models (LLMs) has demonstrated promising applications in protein sequence design and structural prediction. Moreover, the availability of large-scale antibody datasets such as the Observed Antibody Space (OAS) database has opened new avenues for the development of LLMs specialized for processing antibody sequences. Among these, RoBERTa has demonstrated improved performance relative to BERT, while maintaining a smaller parameter count (125M) compared to the BERT-based protein model, ProtBERT (420M). This reduced model size enables more efficient deployment in antibody-related applications. However, despite the numerous advantages of the RoBERTa architecture, antibody-specific foundational models built upon it have remained inaccessible to the research community. In this study, we introduce Ab-RoBERTa, a RoBERTa-based antibody-specific LLM, which is publicly available at https://huggingface.co/mogam-ai/Ab-RoBERTa. This resource is intended to support a wide range of antibody-related research applications including paratope prediction or humanness assessment.

## 1 Introduction

With the growing significance of antibody-based therapeutics, antibody engineering plays a crucial role in modern medicine[1]. Antibodies, also known as immunoglobulins (Igs), specifically recognize and bind disease-associated antigens, mediating their therapeutic effects through mechanisms such as neutralization[2,3], target delivery[4,5], receptor agonism[6], and antibody-dependent cell-medicated cytotoxicity (ADCC)[7]. Structurally, antibodies are Y-shaped glycoproteins composed of two heavy chains and two light chains interconnected by disulfide bonds. Each chain contains a variable domain at the N-terminus, which confers antigen-binding specificity due to its high sequence diversity, and constant domain at the C-terminus which mediate effector functions[8].

The variable regions of antibodies are not only responsible for the critical function of antigen recognition and binding, but their unique sequences also reflect the underlying genetic blueprint encoded in the germline Ig genes and provide insights into the history and type of B cells such as naïve and memory B cells. The variable regions are encoded by the recombination of germline gene segments–namely IGHV, IGHD, and IGHJ for the heavy chain, and either IGKV/IGKJ or IGLV/IGLJ for the light chains–during V(D)J recombination. This combinatorial assembly, along with junctional diversity introduced at the recombination sites, establishes the initial diversity of the antibody repertoire expressed by naïve B cells, enabling the immune system to recognize an extensive array of antigens[9–11]. Following antigen exposure, B cells undergo clonal expansion and differentiate into memory B cells, accompanied by somatic hypermutation that enhances antibody affinity. Consequently, antigen specificity is shaped by both the V(D)J recombination process and somatic hypermutations, particularly within the complementarity-determining regions (CDRs) of variable region[12,13]. Analysis of antibody sequences within these variable regions can reveal information regarding target specificity, the origin of the germline gene segments, and the lineage and maturation state of B cells. Such insights are crucial for not only engineering antibody sequences, but also understanding the evolution of specific antibody response, and identifying potential biases within the antibody repertoire.

However, the high variability in both amino acid composition and sequence length within antibody variable regions presents significant challenges for sequence alignments. This diversity often leads to mismatches across sequences, resulting in the over-insertion of gaps and a high computational burden[14]. Recent advances in protein large language models (LLMs) have improved the understanding of the contextual relationships between amino acids, facilitating applications such as protein motif identification, as well as structure and function prediction[15–17]. Furthermore, the availability of large-scale antibody variable region sequence dataset, Observed Antibody Space (OAS)[18] has enabled the development of antibody-specific LLMs tailored for these highly diverse sequences.

The OAS[18] database compiles billions of variable heavy (VH) and light (VL) chain sequences from more than 80 different studies, enabling large-scale analyses, including applications of large language models. The utility is further enhanced by the extensive annotation of sequences such as species of origin, B cell sources and subtypes, immune state (e.g. disease or vaccination), and antibody isotypes. Antibody-specific LLMs, pre-trained on OAS database, have been applied to various tasks such as assessing humanness[19], completing partial sequences[20–22], predicting paratopes[23], and modeling antibody structures[24]. Despite their broad applicability, the model weights of these pre-trained LLMs are not fully accessible. Recently, IgBERT and IgT5, two antibody-focused language models built upon the **B**idirectional **E**ncoder **R**epresentation from **T**ransformers (BERT)[25] and **T**ext-**T**o-**T**ext **T**ransfer **T**ransformer (T5)[26] transformer architecture, have been released for public access through HuggingFace [27]. In contrast, models based on the **R**obustly **o**ptimized **BERT** **a**pproach (RoBERTa)[28] offer certain advantages in performance, but its antibody-specific foundation model parameters remain readily inaccessible.

While RoBERTa builds upon the BERT model, it achieves enhanced performance through key modification to the pretraining strategy. Notably, RoBERTa replaces BERT's static token masking with dynamic masking and eliminates the next sentence prediction (NSP) objective. In BERT, the masked language modeling (MLM) task uses a fixed set of masked tokens generated during preprocessing, which remain unchanged throughout training. Conversely, RoBERTa introduces dynamic masking, in which the token masking pattern is randomly re-sampled each time an input sequence is processed. This approach allows the model to encounter diverse masking configurations for the same input, encouraging it to develop more generalized and context-independent representations. Furthermore, the NSP component is removed entirely in RoBERTa, based on findings that its exclusion either maintains or slightly enhances performance on downstream tasks, thereby streamlining the pretraining objective to focus exclusively on MLM[28]. Among antibody-specific LLMs, AntiBERTa[23], AbLang[21], and Sapiens[19] are built upon the RoBERTa architecture. However, the foundational (pretrained) model weights for AntiBERTa have not been made publicly available. Similarly, for AbLang and Sapiens, only task-specific fine-tuned model parameters are accessible, while their original pretrained weights remain unreleased.

In our study, we trained RoBERTa architecture with 402 milion antibody sequences from the OAS database. As a result, our antibody sequence foundational model, named as Ab-RoBERTa, is accessible at https://huggingface.co/mogam-ai/Ab-RoBERTa. Additionally, our findings revealed that the single amino acid (SAA) tokenizer outperformed both the double amino acid (DAA) and byte pair encoding (BPE)[29] tokenizers. In downstream tasks, including targeted antigen and B cell type predictions, Ab-RoBERTa consistently achieved remarkable task-specific performance, ranking just after IgT5, while only consuming approximately five-fold less time for fine-tuning. These results indicate that Ab-RoBERTa delivers competitive performance in downstream applications while offering significant gains in computational efficiency. Taken together, Ab-RoBERTa demonstrates strong potential for broad and general applicability in antibody research.

## 2 Method
### 2.1 Data preparation

The OAS database[18] offers a comprehensive and curated repository of annotated antibody sequences derived from a wide range of publicly available B-cell receptor (BCR) repertoire sequencing studies. From this resource, we acquired 2 billion antibody sequences encompassing the variable regions of heavy and light chains. In this study, we focused exclusively on human antibody sequences and applied additional filtering criteria proposed by Leem et al[23]. Specifically, sequences were excluded if the framework 1 (FR1) region was shorter than 20 amino acids or the framework 4 (FR4) region was shorter than 10 amino acids. Overall, a total of 574 million sequences were obtained and randomly spitted into training, testing, and validation set, comprising 402 milion, 86 milion, 86 milion sequences, respectively.

### 2.2 Tokenization

We investigated minimal token representations that best capture the functional characteristics of antibody sequences. The SAA tokenizer comprises the 20 standards amino acid residues along with five special tokens commonly employed in LLM training–namely, start, end, padding, unknown, and mask tokens. Each amino acid is assigned a unique token, consistent with standard practices in protein and antibody language modeling. In addition, we implemented a DAA tokenizer, which builds on the SAA tokenizer vocabulary by treating every pair of consecutive amino acid as a discrete token. This result in a total vocabulary of 425 tokens, including the five special tokens, 20 single amino acid tokens and 400 possible dipeptide combinations. Finally, we tested a BPE[29] tokenizer, trained on unpaired heavy and light chain antibody sequence datasets. This approach yielded a vocabulary consisting of 10,260 tokens, determined through the application of the BPE tokenizer algorithm.

### 2.3 Pre-training

Ab-RoBERTa is implemented using RoBERTa[28] architecture, an advanced transformer-based model derived from BERT[25] with several architectural and training enhancements. Unlike

BERT, RoBERTa incorporates dynamic token masking, eliminates the NSP objective, and employs refined hyperparameter settings, resulting in improved semantic representation and achieved strong performance across a range of natural language processing tasks[28]. For this work, we adopted the default configuration specified in *RobertaConfig* class from HuggingFace Transformers library, except for two modifications: the maximum sequence length (max_position_embeddings) was set to 150 to minimize padding and the vocabulary size (vocab_size) set to 25, 425, and 10,260 for SAA, DAA, BPE, respectively. (**Table 1**).

Table 1. Hyperparameters for Ab-RoBERTa configuration

| Hyperparameters | Value |
|---|---|
| vocab_size | ( 25, 425, 10260 ) (SAA, DAA, BPE) |
| max_position_embeddings | 150 |
| hidden_size | 768 |
| num_hidden_layers | 12 |
| num_attention_heads | 12 |
| intermediate_size | 3,072 |
| hidden_act | gelu |
| hidden_dropout_prob | 0.1 |
| attention_probs_dropout_prob | 0.1 |
| max_position_embeddings | 152 |
| layer_norm_eps | 1e-12 |
| position_embedding_type | absolute |

We pre-trained Ab-RoBERTa using the MLM objective. It involves randomly masking a subset of tokens within an input sequence and training the model to predict these masked tokens based on their surrounding context. Following the RoBERTa masking strategy, 15% of the tokens in each sequence were selected for masking. Of these, 80% were substituted with a special [MASK] token, 10% were replaced with randomly chosen alternative tokens, and the remaining 10% were left unchanged to preserve contextual variability. This strategy enabled the model to learn contextual representations of amino acid sequences in antibody, allowing it to capture the underlying biochemical relationships and dependencies.

The model was trained on three NVIDIA A100 10GB GPUs with a batch size of 384 per device. We employed the AdamW optimizer with a weight decay coefficient of 0.01, an epsilon value of 1e-6, and beta2 parameter set to 0.98. A linear learning rate scheduler was used, configured with an initial rate of 1e-4 and a warmup_step of 30,000. The training was conducted over 6 epochs and completed in approximately 654 hours.

**2.4 Fine-tuning**

To evaluate the utility of Ab-RoBERTa across a range of antibody-related downstream tasks, we compiled a selection of publicly available antibody and protein LLM and conducted a comparative analysis through fine-tuning. Specifically, we included IgBERT[27], IgT5[27], AntiBERTy[20], ProtBERT[16], and ProtT5[16] alongside Ab-RoBERTa for our benchmarking experiments. Two models, Ablang[21] and Sapiens[19], were excluded due to technical incompatibilities. For Ablang, we identified a mismatch between the model architecture and

tokenizer–utilizing a RoBERTa-based architecture in combination with a BERT-based tokenizer–when retrieved from HuggingFace. In the case of Sapiens, while all other models were accessible through HuggingFace, its implementation depends on fair-seq-based code, which diverge from the setup used for the other models and posed integration challenges.

All six LLMs were fine-tuned under identical hyperparameter settings, as detailed in **Table 2** and evaluated across three downstream classification tasks, summarized in **Table 3**. To ensure the robustness and reliability of the results, each model was fine-tuned across five different random seed initializations, and final performance was averaged over these runs. The fine-tuning was performed on NVIDIA A100 GPUs. To accommodate model-specific memory requirements while maintaining a fixed batch size, we utilized A100 GPUs with different memory capabilities: the T5-based models (IgT5 and ProtT5) were fine-tuned on 80GB A100 GPUs, while the remaining models were fine-tuned using 40GB A100 GPUs.

**Table 2**. **Hyperparameters for the finetuning**

| Hyperparameters | Value |
|---|---|
| per_device_train_batch_size | 16 |
| per_device_eval_batch_size | 16 |
| num_train_epochs | 20 |
| adam_beta2 | 0.99 |
| adam_epsilon | 1e-16 |
| lr_scheduler_type | linear |
| weight_decay | 5e-3 |
| warmup_steps | 100 |
| learning_rate | 1e-5 |

**Table 3. Fine-tuning data**

| Data type | The number of training data | The number of class |
|---|---|---|
| Tageted antigen classification | 20,000 | 5 |
| B cell type classification | 20,000 | 4 |
| Germline V gene classification for heavy chain | 20,000 | 7 |
| Germline V gene classification for light chain | 50,000 | 16 |

## 2.5 Evaluation

Fine-tuned downstream tasks were evaluated in terms of area under the receiver operating characteristic curve (AUROC), accuracy (ACC), F1 score, precision, and recall via Hugging face evaluate-matric (https://huggingface.co/evaluate-metric, version: 0.4.3).

### 2.5.1 Area under the receiver operating characteristic curve (AUROC)

AUROC quantifies a model's ability to discriminate between positive and negative classes across all possible classification thresholds. A higher AUROC value, approaching 1, reflects superior discriminatory performance. For instance, an AUROC of 1 denotes a perfect classifier, 0.5 corresponds to performance equivalent to random guessing, and values below 0.5 suggest performance worse than random—often due to model deficiencies or labeling errors. To compute

AUROC in the multi-class setting, we employed the one-vs-rest (OvR) strategy: each class was individually treated as the positive class while the remaining classes were considered negative. The resulting binary AUROC scores were then averaged to yield the final metric.

$$AUROC = \frac{\sum_{p \in positive} \sum_{n \in negative} \mathbb{I}(score_p > score_n) + \frac{1}{2}\sum_{p \in positive}\sum_{n \in negative}\mathbb{I}(score_p = score_n)}{|positive| \times |negative|}$$

Here, $\mathbb{I}(condition)$ denotes an indicator function that return 1 when the condition is true, and 0 otherwise, $score_p$ and $score_n$ represent the predicted score for positive and negative instances, and $|positive|$ and $|negative|$ refer to the total number of positive and negative samples, respectively. The unweighted average AUROC across k classes in a multi–class setting is calculated as:

$$AUROC_{multi-class} = \frac{1}{k}\sum_{k=1}^{k} AUROC_k$$

### 2.5.2 Accuracy (ACC)

Accuracy quantifies the ratio of correctly predicted instances to the total number of instances, serving as an overall measure of correctness of the classifier. While a higher accuracy typically reflects stronger model performance, it may be deceptive in scenarios involving class imbalance.

$$Accuracy = \frac{TP + TN}{TP + FP + FN + TN}$$

where, TP denotes true positives, TN is true negatives, FP is false positives, and FN is false negative.

### 2.5.3 F1 score, precision, and recall

The F1 score offers a balanced measure of a classification model's precision and recall by computing their harmonic mean. This metric is especially valuable in contexts with imbalanced class distributions, where accuracy may not provide a reliable performance measure. Precision refers to the fraction of true positive predictions among all instances labeled as positive by the model, while recall denotes the fraction of true positive cases correctly identified among all actual positive instances. Individual scores were subsequently averaged to obtain the final performance metric.

$$F\ score = 2 \times \frac{Precision \times Recall}{Precision + Recall}$$

$$Precision = \frac{TP}{TP + FP}$$

$$Recall = \frac{TP}{TP + FN}$$

## 3 Result
### 3.1 Effect of different tokenization on antibody sequences

Tokenization refers to the transformation of input data into an ordered set of discrete units, known as tokens, which represents words, subwords, or characters. These tokens serve as the fundamental elements that a LLM can interpret and process. Each token is mapped to a unique integer identifier and subsequently converted into a vector representation (embedding vector) through the LLM's learned parameters. In proteins, the primary structural unit is a sequence composed of 20 distinct amino acid types. Previous studies involving protein-based LLMs have typically treated individual amino acids as the fundamental token units[16].

To date, alternative tokenization strategies beyond single amino acid units have received limited attention in protein LLM research. In this study, we systematically evaluated three distinct tokenization schemes to determine the most effective discrete representation for antibody-specific language modeling. Alongside the standard SAA tokenizer, we explored a DAA tokenizer and BPE[29] tokenizer. Given that amino acid properties are often influenced by adjacent residues, capturing dipeptide-level information may enhance the contextual representation of sequences. In case of BPE, the algorithm offers a data-driven approach by iteratively merging frequently occurring character sequences into subword units, enabling the formation of more informative and compact token representations. Additionally, these multi-residue token units contribute to sequence length reduction after tokenization, potentially improving computational efficiency.

To assess the representational capacity of each tokenization method, we trained models on 40 million antibody sequences using the RoBERTa architecture with SAA, DAA, and BPE tokenizers, respectively. Subsequently, we randomly selected 3,000 antibody heavy sequences from the test dataset, transformed them into embedding vectors, and visualized the representations using uniform manifold approximation and projection (UMAP)[30]. The results demonstrated that embeddings derived from both SAA and BPE tokenizers were capable of distinguishing seven distinct germline V gene families. Notably, only the SAA tokenizer-based embeddings clearly separated B cell subtypes (naïve and memory) and target antigen classes (healthy/celiac disease, HIV, and SARS-CoV-2). These findings indicate that SAA tokenizer provides the most robust and informative representation among the evaluated tokenization methods.

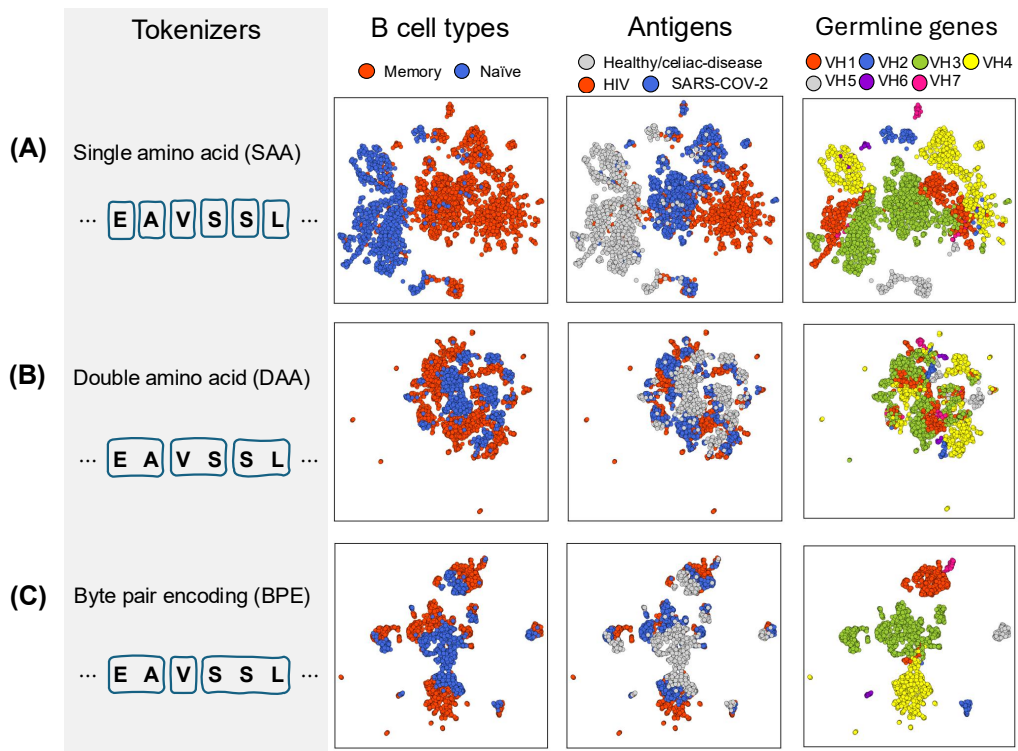

**Figure 1**. **Visualization of pre-trained embeddings generated from three distinct tokenization strategies using UMAP**[30]. Antibody sequences were tokenized using (A) SAA, (B) DAA, and (C) BPE approaches. A subset of 3,000 antibody heavy chain sequences was randomly sampled from the test dataset and transformed into embedding vectors using each tokenizer. In the UMAP plots, each data point represents a single antibody sequence, with colors indicating associated biological attributes.

## 3.2 Insights derived from the embeddings of pre-trained LLMs

Based on the observations above, we developed an antibody-specific large language model, referred to as Ab-RoBERTa, utilizing the SAA tokenizer and trained on 402 million antibody sequences from OAS database—including both unpaired heavy and light chains—using the RoBERTa architecture. To evaluate the model's ability to infer biological attributes solely from sequence information, we compared Ab-RoBERTa against five additional models, including a one-hot encoding baseline as a negative control. All five models employed the SAA tokenizer. Among them, the antibody-specific models—AntiBERTy, IgBERT, and IgT5—were trained on sequences sourced from the OAS database. In contrast, the two general protein language models, ProtBERT and ProtT5, were trained on UniRef100 and UniRef50 datasets [31], respectively.

A subset of 3,000 antibody heavy chain sequences was randomly selected from the test dataset, transformed into embedding vectors using each of the six LLMs and the one-hot encoding baseline, and subsequently visualized using UMAP. The one-hot encoding approach resulted in overlapping data points, failing to distinguish between B cell subtypes or target antigen classes, though it did show clear clustering by germline V genes. Similar patterns were observed for ProtT5, ProtBERT, and AntiBERTy, which displayed consistent germline-based clustering with limited separation in the other categories. In contrast, Ab-RoBERTa, IgBERT, and IgT5 demonstrated more robust clustering across all three biological attributes—germline V genes, B cell types, and target antigens—indicating a stronger capacity for capturing diverse sequence-level biological signals.

Model size, often quantified by the number of parameters, is generally associated with enhanced performance in downstream tasks [32]. However, this benefit comes with a significant trade-off in terms of training cost. Larger models demand greater computational resources including time, hardware, and energy, which can affect their practical deployment, particularly in latency-sensitive or resource-limited environments. In our comparison, the T5-based models ProtT5 and IgT5 each comprise approximately 3 billion parameters. The BERT-based models, ProtBERT and IgBERT, contain around 420 million parameters, while AntiBERTy, though also BERT-based, is significantly smaller with 26 million parameters due to architectural modifications. Our RoBERTa-based model, Ab-RoBERTa, has 125 million parameters (**Figure 2b**). The following section will examine the trade-off between fine-tuning performance and training efficiency across these models.

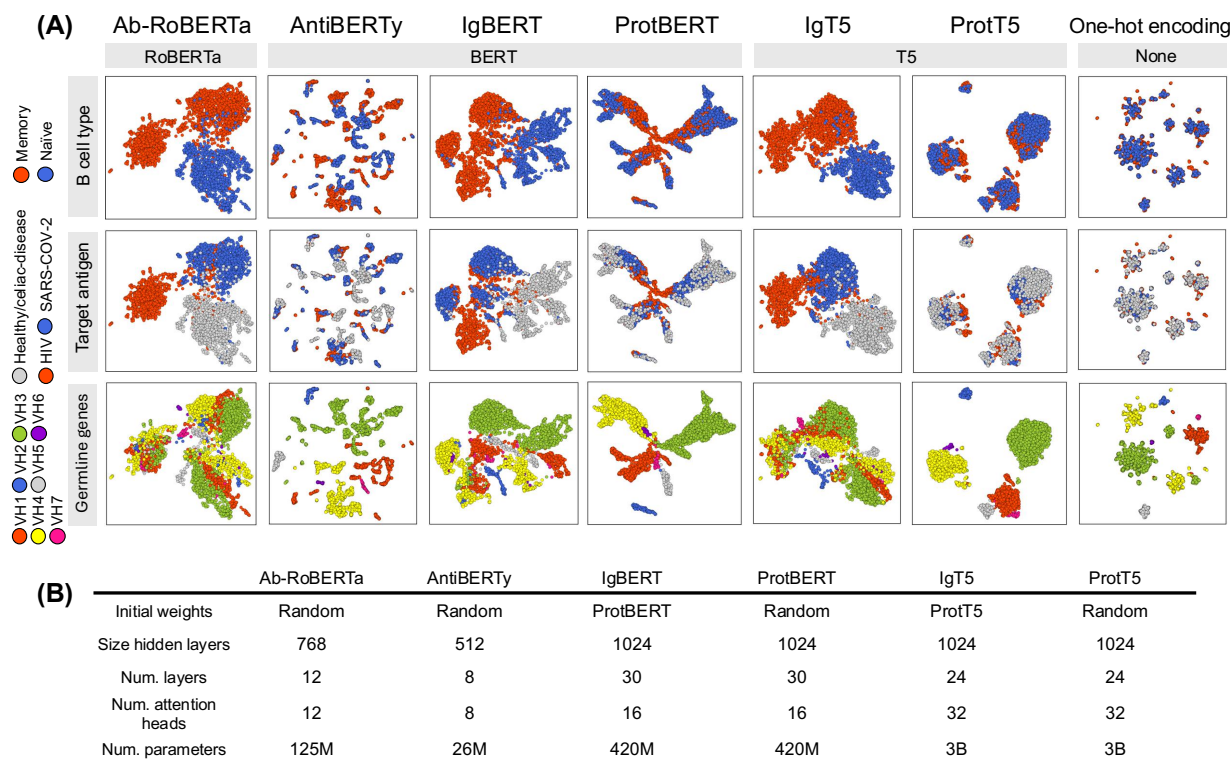

**Figure 2. LLM comparison using pre-trained embedding representations.** (A) A subset of 3,000 antibody heavy chain sequences was randomly selected and individually encoded into embedding vectors using seven distinct models. The resulting embeddings were visualized using UMAP[30], where each point corresponds to a single antibody sequence, and color annotations reflect relevant biological features. (B) Model configurations, including the architecture and parameter size, are summarized.

### 3.3 Comparative evaluation of fine-tuning performance across LLMs

We evaluated the performance of six LLMs by adding a classification head to their final layer and assessing their predictive capabilities using AUROC, ACC, F1 score, precision, and recall. The evaluation was conducted across three classification tasks: (1) classification of five target antigen categories—HIV, SARS-CoV-2, muscle-specific tyrosine kinase (MuSK) myasthenia gravis, acetylcholine receptor antibody-positive (AChR) myasthenia gravis, and cytomegalovirus—for both heavy and light chains; (2) classification of B cell subtypes, including four types for heavy chains (naïve B cells, memory B cells, plasmablasts, and germinal center B cells) and three types for light chains (naïve B cells, memory B cells, and plasmablasts); and (3) classification of germline V gene families, comprising seven types (VH1–VH7) for heavy chains and sixteen types (VK1–VK6 and VL1–VL10) for light chains (**Table 3**).

In the targeted antigen classification task, IgT5 demonstrated the highest performance for heavy chain sequences, with Ab-RoBERTa ranking closely behind. For light chains, performance varied slightly, with IgT5 and Ab-RoBERTa exhibiting comparable results (**Table 4**). Similarly, in the B cell type classification task, both of IgT5 and Ab-RoBERTa performed at a similar level; however, Ab-RoBERTa showed a slight advantage on heavy chain sequences, while IgT5 outperformed on light chains (**Table 5**). Germline V gene prediction was found to be a relatively straightforward task, as all models reached perfect accuracy after just one epoch, with all evaluation metrics reaching a value of 1.0. Given this immediate convergence, further fine-tuning was not pursued. When considering the embeddings from one-hot encoding, these results indicate that germline V gene classification is likely driven predominantly by primary sequence similarity, rather than by higher-order contextual or semantic features captured by pretrained models.

From the perspective of training efficiency, we evaluated both the rate at which the training loss converged to zero and the total training time over 20 epochs. While some variability was observed across tasks and random seeds, IgT5 consistently demonstrated the fastest loss convergence to zero, followed by Ab-RoBERTa (**Figure 3a**). Training duration generally scaled with model size: T5-based models, being the largest, required approximately 16 hours. ProtBERT and IgBERT completed training in about 7 and 5 hours, respectively. Ab-RoBERTa required roughly 3 hours, whereas AntiBERTy, the smallest model size, completed training in just 2 hours (**Figure 3b**).

**Table 4. Targeted antigen classification evaluation**
The row corresponding to Ab-RoBERTa was shaded in gray for visual distinction. For each evaluation metric, the highest value is indicated in red, while the second-highest value is showed in bold. All evaluations were conducted over five independent runs, with results reported as the mean and the standard deviation.

| Model | Chain | AUROC | ACC | F1 | Precision | Recall |
|---|---|---|---|---|---|---|
| Ab-RoBERTa | heavy | **0.850**±0.006 | **0.551**±0.013 | **0.546**±0.019 | **0.562**±0.011 | **0.551**±0.013 |
| AntiBERTy | heavy | 0.837±0.004 | 0.539±0.010 | 0.537±0.011 | 0.545±0.010 | 0.539±0.010 |
| IgBert | heavy | 0.823±0.009 | 0.511±0.019 | 0.509±0.021 | 0.527±0.020 | 0.511±0.019 |
| IgT5 | heavy | 0.858±0.007 | 0.562±0.017 | 0.556±0.022 | 0.583±0.017 | 0.562±0.017 |
| ProtBert | heavy | 0.786±0.009 | 0.450±0.010 | 0.439±0.011 | 0.459±0.011 | 0.450±0.010 |
| ProtT5 | heavy | 0.819±0.005 | 0.496±0.013 | 0.487±0.020 | 0.522±0.009 | 0.496±0.013 |
| Ab-RoBERTa | light | **0.830**±0.002 | 0.524±0.005 | 0.515±0.006 | **0.526**±0.008 | 0.524±0.005 |
| AntiBERTy | light | 0.817±0.003 | 0.504±0.009 | **0.501**±0.007 | 0.511±0.007 | 0.504±0.009 |
| IgBert | light | 0.801±0.003 | 0.469±0.009 | 0.464±0.007 | 0.486±0.007 | 0.469±0.009 |
| IgT5 | light | 0.832±0.005 | **0.520**±0.016 | 0.498±0.030 | 0.528±0.013 | **0.520**±0.016 |
| ProtBert | light | 0.790±0.004 | 0.461±0.011 | 0.453±0.012 | 0.468±0.012 | 0.461±0.011 |
| ProtT5 | light | 0.798±0.007 | 0.472±0.012 | 0.471±0.014 | 0.493±0.010 | 0.472±0.012 |

**Table 5. B cell type classification evaluation**
The row corresponding to Ab-RoBERTa was shaded in gray for visual distinction. For each evaluation metric, the highest value is indicated in red, while the second-highest value is showed in bold. All evaluations were conducted over five independent runs, with results reported as the mean and the standard deviation.

| Model | Chain | AUROC | ACC | F1 | Precision | Recall |
|---|---|---|---|---|---|---|
| Ab-RoBERTa | heavy | **0.890**±0.002 | **0.668**±0.007 | **0.669**±0.009 | **0.678**±0.007 | **0.668**±0.007 |
| AntiBERTy | heavy | 0.851±0.006 | 0.619±0.013 | 0.621±0.013 | 0.628±0.014 | 0.619±0.013 |
| IgBert | heavy | 0.840±0.009 | 0.614±0.006 | 0.609±0.013 | 0.621±0.013 | 0.614±0.006 |
| IgT5 | heavy | **0.883**±0.007 | **0.666**±0.010 | **0.667**±0.011 | **0.678**±0.015 | **0.666**±0.010 |
| ProtBert | heavy | 0.821±0.006 | 0.582±0.008 | 0.568±0.012 | 0.583±0.009 | 0.582±0.008 |
| ProTt5 | heavy | 0.841±0.008 | 0.606±0.013 | 0.608±0.013 | 0.615±0.016 | 0.606±0.013 |
| Ab-RoBERTa | light | **0.857**±0.002 | **0.694**±0.005 | **0.694**±0.004 | **0.696**±0.006 | **0.694**±0.005 |
| AntiBERTy | light | 0.833±0.001 | 0.665±0.004 | 0.665±0.004 | 0.667±0.004 | 0.665±0.004 |
| IgBert | light | 0.815±0.007 | 0.640±0.010 | 0.639±0.010 | 0.642±0.011 | 0.640±0.010 |
| IgT5 | light | **0.858**±0.003 | **0.703**±0.006 | **0.701**±0.006 | **0.705**±0.007 | **0.703**±0.006 |
| ProtBert | light | 0.780±0.008 | 0.599±0.011 | 0.598±0.011 | 0.600±0.010 | 0.599±0.011 |
| ProtT5 | light | 0.808±0.004 | 0.629±0.009 | 0.627±0.008 | 0.631±0.009 | 0.629±0.009 |

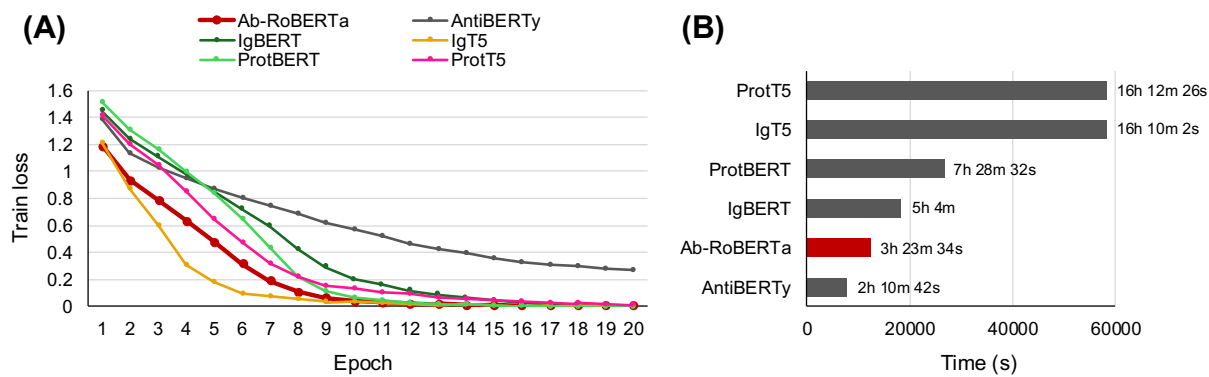

**Figure 3. Evaluation of training efficiency based on training loss progression and computational time.** For the B cell classification task, data from 20 training epochs using a single random seed were shown. (A) Cross-entropy training loss was recorded at each epoch and plotted for six different LLMs to visualize loss trajectories. (B) The cumulative training time required to complete the 20 epochs was presented for each model.

## 4 Conclusion

In this study, we evaluated the performance differences among three distinct tokenization strategies: SAA, DAA, and BPE. Among these, the SAA tokenizer demonstrated superior capability in capturing biologically relevant features of antibody sequences, aligning with observations reported in other protein LLMs. Utilizing the SAA tokenizer, we developed an antibody-specific LLM—referred to as Ab-RoBERTa—based on the RoBERTa architecture and trained on the OAS dataset. Despite its relatively smaller model size compared to BERT- and T5-based language model, Ab-RoBERTa exhibited strong performance, closely following IgT5, which achieved the highest metrics across most evaluation criteria. In addition, both IgT5 and Ab-RoBERTa demonstrated rapid convergence to zero training loss. Notably, Ab-RoBERTa offered significant advantages in computational efficiency, with fine-tuning requiring only one-fifth of the time needed for IgT5. Given its training efficiency, Ab-RoBERTa offers a practical solution for deployment in environments with limited computational resources or strict latency constraints. To support the wider antibody research community, the Ab-RoBERTa model is publicly released, enabling its application across a broad spectrum of antibody-related studies.

## 5 Availability

Ab-RoBERTa is accessible at https://huggingface.co/mogam-ai/Ab-RoBERTa.